\documentclass[letterpaper,twocolumn,fleqn]{article} 

\usepackage{ist}
\usepackage{caption}
\usepackage{subcaption}
\usepackage{booktabs}
\usepackage{hyperref}
\usepackage[table]{xcolor}
\usepackage{tabu}
\usepackage{tabularx}

\title{Using a GAN to Generate Adversarial Examples to Facial Image Recognition}

\author{Andrew Merrigan~$^1$ and Alan F. Smeaton~$^{1,2}$\\
$^1$School of Computing and~ $^2$Insight Centre for Data Analytics\\
Dublin City University, Glasnevin, Dublin, 9, Ireland.\\
Email: alan.smeaton@dcu.ie}

\date{} 

\begin{document} 

\maketitle 

\thispagestyle{empty} 


\begin{abstract}
Images posted online present a privacy concern in that they may be used as reference examples for a facial recognition system.  Such abuse of images is in violation of  privacy rights but  is  difficult to counter. It is well established that adversarial example images can be created for recognition systems which are based on deep neural networks. These adversarial examples can be used to disrupt the utility of the images as reference examples or training data.  In this work we use a Generative Adversarial Network (GAN) to create adversarial examples to deceive facial recognition and we achieve an acceptable success rate in fooling the face recognition. Our results reduce the training time for the GAN by removing the discriminator component. Furthermore, our results show knowledge distillation can be employed to drastically reduce the size of the resulting model without impacting performance indicating that our contribution could run comfortably on a smartphone.\footnote{(c) Copyright of this article remains with the Society for Imaging Science and Technology (IS\&T), 2022}
\end{abstract}

\section{Introduction and Problem Statement}
Deep neural networks (DNNs) have superior performance to more conventional approaches when applied to a variety of computer vision tasks including object recognition, activity detection and facial recognition. DNN’s are however susceptible to adversarial examples which are images crafted so that the machine's interpretation differs greatly from that of a person looking at the same image \cite{brown2018adversarial}.
Several recent works have shown that such adversarial examples are actually effective against a variety of trained models, not just those  they have been generated to mislead \cite{brown2018adversarial}.

The use of facial recognition systems (FRS) coupled with  unauthorised use of images taken from social media  is a threat to individuals' privacy. The use of facial recognition systems is  increasing worldwide. The  development of open-source facial recognition models based on DNNs has made it possible for anyone with access to moderate computing power to create an FRS. More troubling  is  corporate use of  images from social media such as  Clearview.ai who have scraped approximately 3 billion images for an FSR without  individuals’ consent \cite{AdversarialExamplesOpportunitiesChallenges8842604}.

The work reported in this paper  examines the feasibility of employing the AdvGAN++ architecture to generate adversarial examples targeted against facial recognition. This design was proposed and tested in the generation of adversarial examples using the MNIST and CIFAR-10 datasets. As it has yet to be examined in the context of facial recognition its performance characteristics in this task are unknown.

Furthermore, the actual utility of discriminators in the generation of adversarial image examples via a GAN will be examined by determining the discriminator's function within the network, ultimately with the possibility of its removal. Removing the discriminator from the network would provide a significant reduction in the computational cost of training the network. It would also reduce the complexity of the model making the training of the model simpler by removing many of the hyper-parameters.  

In summary, the motivation behind this work is to create a small effective model, which can be deployed on user devices to enable them to convert images into adversarial examples, prior to putting them online and  the novelty and contribution of the paper is the reduction and simplification of the process of generating adversarial faces. 
The rest of this paper is organised as follows. In the next section, we describe some  related work, followed by a section on the datasets we used. The subsequent section describes our experiments, covering model design, performance, the removal of the discriminator from the GAN and the distillation of the generator’s knowledge.

\section{Related Work}

There are many  approaches to facial recognition however the general pipeline consists of four stages namely detecting, alignment, representation and verification \cite{9259802}.
Through the use of deep convolutional neural networks (DCNNs), facial recognition systems have reached a level of human-like performance including on the set of ``Labeled Faces in the Wild'' (LFW) dataset. However, problems still exist because of the high degree of variability among features such as head position, lighting, facial expression and ageing. These challenges cumulatively have been termed the problem of PIE (pose-illumination-expression) in face recognition.
The detection stage is no longer considered a challenging task for frontal faces.

Research into adversarial examples in facial recognition typically targets the representation stage which is responsible for creating a feature vector to represent a given face in an image. 
There are a  number of methods for generating adversarial examples which either take an optimisation approach or use a neural network to generate a perturbation.
In optimisation based methods the parameters of the model are held constant while an algorithm attempts to find changes (or perturbations) that can be made to the input image to achieve the desired output. This is done either within the constraints that the perturbation is less than some maximum or within some bounded area of the input. Many different approaches have been suggested to solve the optimisation problem such as gradient descent \cite{Fendley2020JacksofAll}, FGSM \cite{papernot2015limitations}, L-BFGS \cite{szegedy2014intriguing} and Projected Gradient Descent \cite{madry2019deep}. 

Originally the emergence of adversarial examples was suggested to originate in the non-linearity of deep neural networks or in insufficient regularisation being applied to a model. However, it can be shown that adversarial examples arise from a model being too linear.  It has also been demonstrated that training using adversarial examples with correct labels termed adversarial training, can be used to provide regularisation benefits \cite{goodfellow2014explaining}.
Several papers have shown that these adversarial examples are transferable. This means that adversarial examples are not limited to specific DCNN models or the models they have been trained against. This makes it unnecessary to know exactly what models the adversarial example will be used against. The transferability will be maximised between models with similar architecture, number of parameters and with high test accuracy \cite{kurakin2017adversarial}.

Broadly, the research into adversarial examples can be placed into two different categories. One category calculates a minimal set of pixel value changes based on a specific input image which is the approach we take here. The benefit of this approach is that the original and modified images are usually indistinguishable from each other. In some literature this approach of bespoke modification of an input image is termed ``cloaking''. The effectiveness of a cloaking method is not just measured by  the proportion of test samples which are incorrectly identified but also by how visually similar the cloaked image is to the original image.
The primary disadvantage of this approach is the computational cost of cloaking an input image. To cloak a single image takes approximately 1 minute on a modern CPU powered computer \cite{shan2020fawkes}.

There are several methods which can be used to assess the similarity of two images from simple approaches that only consider the mean squared error of the averaged pixel intensities, 
to methods which consider the capabilities of human vision \cite{SSIM1284395}. The approach found in the literature is a typically structural similarity index measure (SSIM) or a derived value such as DSSIM, which is a measure of the structural dissimilarity \cite{shan2020fawkes}.
Since SSIM is a full reference metric both the original and distorted images are required to compute the value. SSIM compares local patterns within the image after  pixel values have been normalised for luminance and contrast \cite{SSIM1284395}.

Generating adversarial examples using a neural network  takes considerably less time  when compared with optimisation approaches. \cite{bai2021aigan} found that When using a neural network to generate adversarial examples the resulting training process is similar to that of a Generative Adversarial Network (GAN) where the goal is to find a weakness in another model.
The authors in \cite{baluja2017adversarial} propose Adversarial Transformation Networks (ATNs). ATNs come in two distinct types, one model type which is based on a CNN termed a Perturbation ATN (P-ATN) and another based on an autoencoder called an Adversarial Autoencoding (AAE).
AAE may have some issues in terms of detectability in that they remove the high frequency data from the input image due to limitations in the architecture \cite{baluja2017adversarial}.

Some researchers have used modified GANs to create adversarial examples and examples include AdvGAN \cite{xiao2019generating}, SLP-GAN \cite{SLP-GAN_10.1007/978-3-030-65299-9_1},
 and Attack-Inspired GAN (AI-GAN) \cite{bai2021aigan}.

AdvGAN has a generator which attempts to create a perturbation for an input image to fool two separate adversaries. One of the adversaries attempts to find the modified image when given both the original and perturbed image. The other adversary is the network that is being targeted \cite{xiao2019generating}.
The performance and training time of AdvGAN was later improved by using a feature space representation of the input image to train the generator rather than the whole input image \cite{jandial2019advgan++}.

An adaption of the original AdvGAN termed PcAdvGAN is similar to the improved AdvGAN++ in that it first encodes the input image prior to the generator network. The encoding used however was based on principle component analysis of different segments of the input image. Unlike the other GAN designs this model was tested on facial recognition systems not classification problems \cite{face_segment9106374}.  

AI-GAN is a similar design to that of the first implementation of AdvGAN however it incorporates the desired target class for the input image into the generator, as well as incorporating the target into the loss function. This contrasts with the AdvGAN in which the adversarial example will have a non-specific class selected by the generator \cite{bai2021aigan}.  

\section{Datasets}

In this work the {\it CelebFaces Attributes Dataset } (CelebA) dataset was used as a  training set with the {\it Labeled Faces in the Wild} (LFW) dataset split into a validation and test set \cite{liu2015faceattributes, LFWTech}.
CelebA is a large-scale face attribute dataset containing 202,506 images of 10,177 individuals \cite{liu2015faceattributes}. The LFW dataset contains 13,233 images of 5,749 individuals, it is a publicly available benchmark focused on face verification of subjects in unconstrained environments \cite{LFWTech}.

CelebA  was used for training because whilst the images are typically much more constrained when compared to the LFW, CelebA is much larger. Therefore, the LFW dataset is valuable for testing as it contains images of individuals in more natural unconstrained settings. Using the LFW dataset for testing also allows for comparisons with work by others.

\begin{figure}[ht]
    \centering
    \includegraphics[scale=0.3]{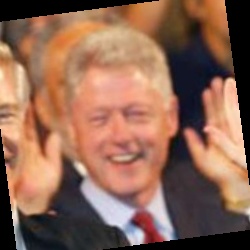}
    \includegraphics[scale=0.3]{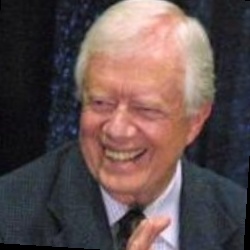}
    \includegraphics[scale=0.3]{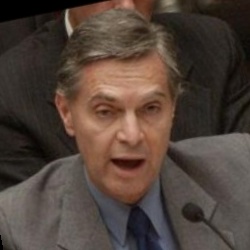}
    \caption{Some original aligned images in the CelebA dataset}
    \label{fig:original_images}
\end{figure}

The images in both datasets are of varying sizes but are all in JPEG format with each accompanied by a unique name for the primary subject of the image, who is featured prominently in the centre with some images also including multiple other individual faces  \cite{LFWTech}. A sample of faces from CelebA is shown in Figure~\ref{fig:original_images}.

The LFW dataset was split into a validation set of 2,677 images and a test of 10,552 images.  Four images of Hollywood actor Nicolas Cage were reserved from the dataset for use as a target image.
This ensured a number of images of each individual were present, which maximises the distance that an image's representation must be moved to be incorrectly classified. 

The images in the LFW dataset do not contain close-cropped images of the subjects' faces that are typically used as an input in face verification tasks. To transform the images into this form a pre-trained CNN model was used to place bounding boxes over each of the faces in an image. The largest of these  was then used to crop the primary subject's face from the original image. Initial attempts to use a HOG model resulted in several images where faces were obscured, being excluded or poorly cropped.
This cropping also removes interference from other individuals whose faces may be present in the same image. It also reduces the variance between images as variations in pose and the image background are largely removed \cite{LFWTech}.
After cropping, all of the images were scaled using bilinear interpolation to 160x160 pixels in size.

A  concern with the dataset is the lack of variance from lighting, pose, and perspective transformation but also the range of characteristics of subjects in the images. The dataset has a disproportionately large number of males compared to females. Many races and ethnicities are not represented highly in the dataset or even at all. The age range of subjects is also restricted with few young or old individuals  \cite{LFWTech}. However, this dataset is used as a benchmark for several facial recognition models which will allow the success of our own created adversarial models to be measured in terms of decreases in accuracy between the original LFW dataset and an adversarial version of the LFW dataset.

\section{Experiments}
\subsection{Model Design}
 
\begin{figure}[ht!]
    \centering
    \includegraphics[width=\linewidth]{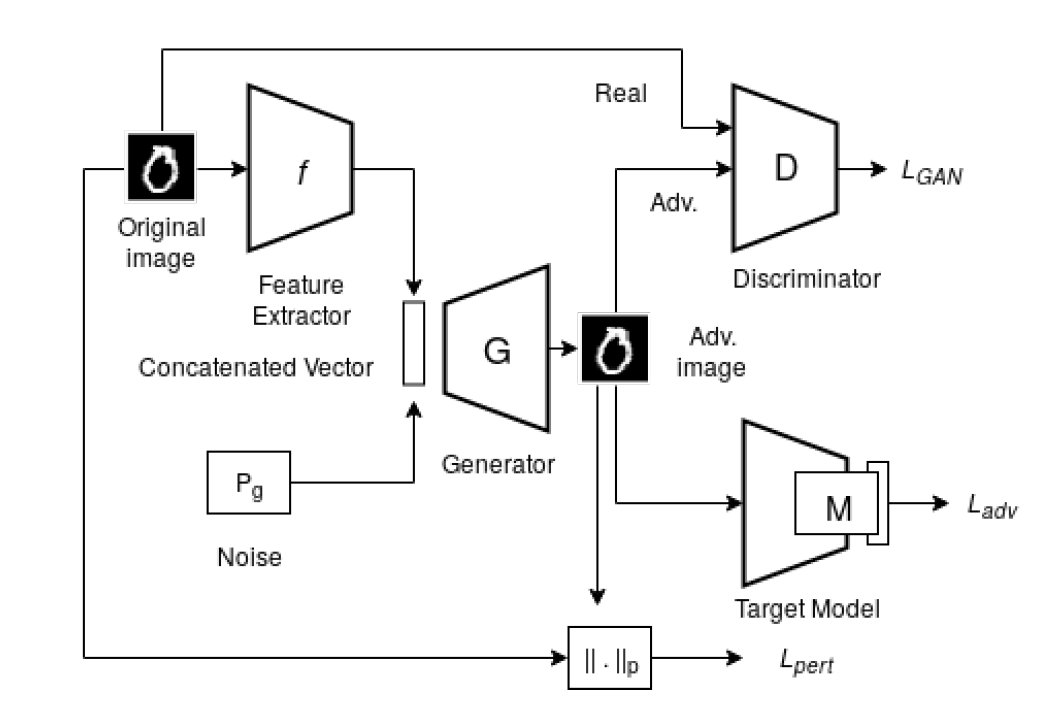}
    \caption{Original AdvGAN++ Architecture taken from \cite{jandial2019advgan++}}
    \label{fig:advgan_arch}
\end{figure}

Initial work was performed on an AdvGAN++ architecture   modified to operate against facial recognition models. The original AdvGAN++ architecture is shown in  Figure~\ref{fig:advgan_arch}. 
 
One of the main differences between the targeted and untargeted tasks is the output of the  model, which for the target facial recognition model used in our experiments was a 128-digit encoding. This required that the loss function had to be adjusted to accommodate this change and the loss function for the generator is formulated as:
 
\[L(G) = \alpha L_{GAN} + \beta L_{adv} + \gamma L_{pert} \]

\noindent 
where $L_{GAN}$ is the binary cross-entropy loss of the discriminator, $L_{adv}$ is a measure of the error introduced into the output of the target model by the changes made by the generator and $L_{pert}$ is a measure of how much the input image was modified. The terms  $\alpha$, $\beta$ and $\gamma$ are hyperparameters used to adjust the relative importance of each term.
The tuning of these three components was difficult and resulted in large   stability problems.

The $L_{adv}$ of the original network was binary cross-entropy which was changed to mean squared error after experimentation with mean squared error, 2-norm and cosine distance. This was used for both the targeted and non-targeted versions of the network. However, in the case of the non-target model of operation the loss function was instead maximised.
 
The $L_{pert}$ term is very important in the functioning of the network. If $L_{pert}$ is more permissive the model changes the structure of the face acting like a style transfer model in the case of a targeted attack or more generally as a form of anonymisation in the case of a non-targeted attack. A restrictive $L_{pert}$ instead results in adversarial noise which is not readily apparent to a human observer.
Experiments were conducted on several different functions for $L_{pert}$. These sought to maintain the total modification of each pixel below a specified threshold. An alternate experiment instead sought to minimise the SSIM between the original and modified image.

\begin{figure}[htb!]
    \centering
    \includegraphics[width=\linewidth]{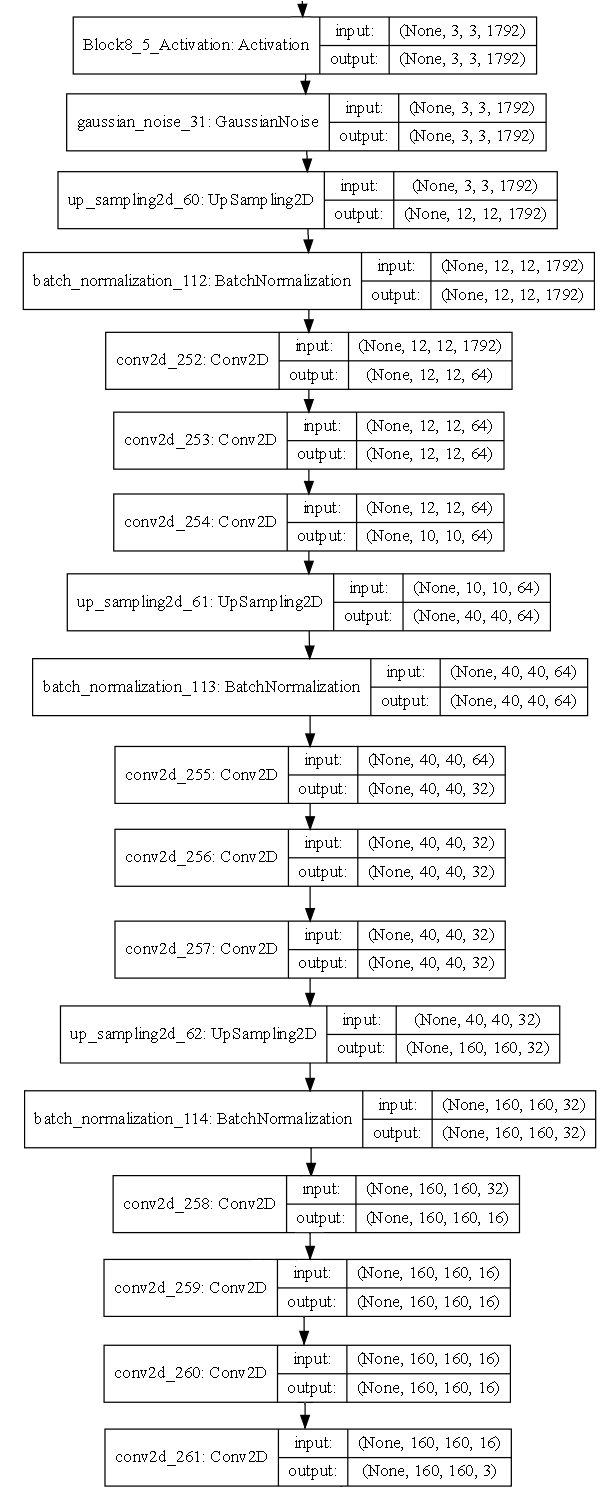}
    \caption{Architecture of generator}
    \label{fig:architecture}
\end{figure}

The structure of the generator is shown in Figure~\ref{fig:architecture} as  a fully convolutional network comprised of up-sampling, convolutional and batch normalisation layers.
It is designed to convert the features of some of the last layers of the target model back into the same dimensions as the original input image. However the pixel values in the outputted perturbation range in value from -1 to +1 so that the model can learn to both increase and decrease each pixel of the input image. The perturbation is then summed with the input image and the result clipped to the range 0 to 1. Originally the increases in the dimensions were achieved using transposed convolutional layers however this led to a tiling effect on the output image. To address this the transposed convolutional layers were replaced with up-sampling. 

The target model used both in  white box experiments and as the feature extraction layers, is an implementation of Facenet which was pre-trained with an accuracy of 0.9963 on the LFW dataset \cite{facenet_Schroff_2015}. Using the feature extraction layers of the target model acts as a type of transfer learning, though existing work does not describe it as such. On the basis that this is a form of transfer learning, experiments added an additional fine-tuning stage to the training.  Consequently some of the higher feature extraction layers of the target model were made trainable and the model was allowed to train again with a lower learning rate.

\subsection{Performance}
\begin{figure}[!htb]
\centering
\includegraphics[width=0.15\textwidth]{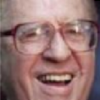}
\includegraphics[width=0.15\textwidth]{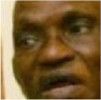}
\includegraphics[width=0.15\textwidth]{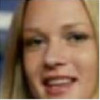}
\includegraphics[width=0.15\textwidth]{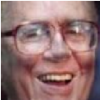}
\includegraphics[width=0.15\textwidth]{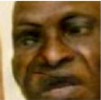}
\includegraphics[width=0.15\textwidth]{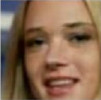}
\caption{Visual artefacts, original (T) and modified (B) images} 
\label{fig:artefacts}
\end{figure}

Basic results for the created models are presented in Table~\ref{table:performance} which includes the introduction of image blurring which is a very basic defence against adversarial examples \cite{gu2015deep}.
The model in several different versions achieves a higher success rate when compared with Fawkes \cite{shan2020fawkes}, a previously proposed iterative cloaking method. 
However, the successful models introduce considerably more graphical changes to the input images than the iterative cloaking method. In some of the generated images the generator acts more as a form of anonymisation than cloaking. Anonymisation whilst destructive is more robust to defensive techniques. These higher-performing models however also tend to introduce more spurious artefacts, examples of which, both generated by the same model are presented in Figure~\ref{fig:artefacts}. 

\begin{table*}[!ht]
\centering
\caption{Performance Characteristics Of Various Models}
\label{table:performance}
\begin{tabular}{p{2.5cm}p{2cm}p{2cm}p{2cm}p{2cm}p{2cm}p{2cm}}
\toprule
\textbf{Model} &
\textbf{Success rate  against white box recogniser }& 
\textbf{Success rate against black box recogniser} &
\textbf{Success rate against white box after blurring} &
\textbf{$L_{pert}$ Threshold used in loss function} &
\textbf{Training Time (mins per Epoch)} &
\textbf{Time to generate  example image (secs)}\\
\midrule 
SSIM & 0.84 & 0.81 & 0.80 & n/a & 20 & 0.024\\ 
\midrule 
High Threshold & 0.97 & 0.94 & 0.96 & 0.2 & 20 & 0.026\\ 
\midrule 
Medium Threshold & 0.98 & 0.99 & 0.87 & 0.1 & 20 & 0.025\\ 
\midrule 
Low Threshold & 0.65 & 0.55 & 0.60 & 0.001 & 20 & 0.025\\ 
\midrule 
Medium Threshold No Target & 0.96 & 0.88 & 0.95 & 0.05 & 20 &  0.026\\ 
\midrule 
Medium Threshold No Discriminator & 0.98 & 0.97 & 0.59 & 0.1 & 15 & 0.024\\ 
\midrule 
Fawkes System  (Shan {\it et al.}, 2020) & 0.74 & 0.97 & 0.54 & n/a & 0 & 8.001\\
\bottomrule
\end{tabular}
\end{table*}

All  models were tested on a black box, a pre-trained implementation of VGGFace\footnote{\url{https://github.com/rcmalli/keras-vggface}}  of which they had no existing knowledge or access. 
The performance of many of the models are maintained when deployed against a blackbox system. However this is contrasted by the performance of the the iterative approach which actually performs better against the blackbox system.

As a GAN based approach uses a fully trained model as opposed to an iterative approach like Fawkes which only makes use of feature extractors from a trained model, the time to generate an adversarial example is far lower. However, the iterative approach does not require any explicit training as opposed to the GAN which requires 20 minutes per epoch.

\subsection{Removal of the Discriminator}
\begin{figure}[htb!]
\centering
\includegraphics[width=0.4\linewidth]{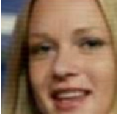}
\includegraphics[width=0.4\linewidth]{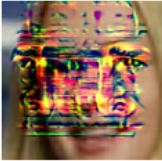}
\caption{Results generated without a discriminator in the GAN, medium threshold (L) and high threshold (R)} 
\label{fig:badNoDisc}
\end{figure}

The GAN used throughout the experiments was retrained without the discriminator both to measure the model’s performance and the computation time required for training.
The success rate of that model  is 0.98 against a white box system. This is comparable to that of a  model with a medium perturbation threshold trained with a discriminator which had a success rate of 0.96 against a white box system. Though the success rate has only been reduced by 0.02 against the white box system, the training time per epoch  reduced from 20  to 15 minutes. This experiment is listed as “Medium Threshold no Discriminator“ in Table~\ref{table:performance}. 
Without a discriminator as part of the architecture, instabilities during training are reduced allowing for higher learning rates. The initial experiments all focused on the use of an Adam optimiser with a learning rate of 1e-4 with beta1 set to 0.5, however without a discriminator the network can be trained with a learning rate of 1e-3 with beta1 at its default of 0.9.
Whilst the performance is comparable with that of the original GAN as the threshold for pixel value modification increases, the results rapidly begin to suffer as shown in Figure~\ref{fig:badNoDisc}.

\subsection{Role of the Discriminator}
Adversarial examples can be deployed as part of a poisoning attack  used to pollute a pool of reference examples used by an FRS  to identify an individual. By inserting adversarial images into this pool genuine images taken from an uncontrolled source such as a surveillance camera will not match reference images. As the  goal in the creation of adversarial examples is to have them inserted into the reference pool it is important that they are not easily detectable as an adversarial example. 

As  illustrated  previously, the same success rate in the creation of adversarial examples may be obtained in a model trained without a discriminator when the level of pixel-level modification is been tightly controlled. We may also find that the discriminator  provides an additional benefit such as preventing the adversarial examples from being easily identifiable. 
If the discriminator was preventing detection then it should be able to distinguish adversarial  from unmodified images but the probability of all  images tested was $\sim 0.5$ regardless of whether they were modified or not. As the discriminator is not learning to detect adversarial images, it is inferred that the discriminator is not teaching the generator to produce adversarial which are more difficult to detect.

\subsection{Model size reduction} 
The created adversarial model contains 426 layers making the values with a total of 22,137,171 parameters with an H5 model size of 85.5MB. To successfully deploy this model in a real-world setting it would ideally have fewer parameters . 
We applied knowledge distillation  to transfer the knowledge gained from an adversarial model which includes the feature extraction layers taken from an FRS model to a much smaller model. Knowledge distillation is a method of model compression that attempts to transfer knowledge from a teacher model to student models while maintaining accuracy \cite{hinton2015distilling}. 

For this experiment, the knowledge learned by the middle threshold model, trained without a discriminator as well as feature extraction layers was transferred to a smaller U-Net architecture \cite{ronneberger2015unet}. 
The resultant distilled model had  a total of 2,058,979 parameters with an H5 model size of 8.04MB. The success rate of the distilled model is 0.98 compared to the teacher model which has a success rate of 0.98. From this,  it is evident that in this case the overall size of the model can be reduced whilst  maintaining the performance.

\subsection{Visualising the Generator Operation}
\begin{figure}[htb!]
    \centering
    \includegraphics[width=0.8\linewidth]{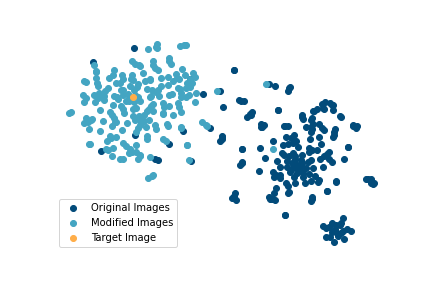}
    \caption{Plot of dimensionally-reduced embeddings using t-SNE for a model with a target}
    \label{fig:tsne}
\end{figure}

One approach taken to visualise the effect of the model was to use  
t-distributed stochastic neighbour embedding (t-SNE) \cite{van2008visualizing}
to reduce the dimensionality of a facial recognition's embedding of modified and original images such that they can be visualised. The results of this are shown in  Figure~\ref{fig:tsne}.

\begin{figure}[htb!]
    \centering
    \includegraphics[width=0.8\linewidth]{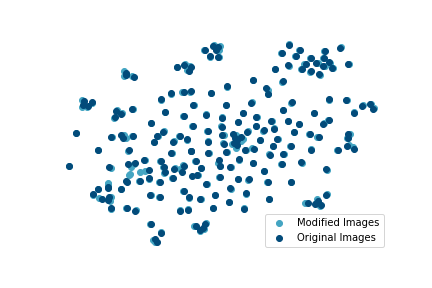}
    \caption{Plot of dimensionally-reduced embeddings using t-SNE for a model without a target}
    \label{fig:no_target_tsne}
\end{figure}
In Figure~\ref{fig:tsne} we  see that as this model is targeted all of the images have been shifted from their original embedding to a  position closer to the target image. This confirms that the original images have been altered, causing their embeddings to form a cluster around the target, far from their original position. This is in contrast to the t-SNE of the untargeted model in which the modifications result in movements in non-specific directions as shown in Figure~\ref{fig:no_target_tsne}.

\section{Conclusions}
The work reported here shows that the AdvGAN++ architecture previously applied to classification on the MNIST and CIFAR-10 datasets can be applied to create adversarial examples for facial recognition. 
The created models have been revealed to have comparable performance to the more traditional iterative approach with a substantially reduced time to generate examples.
It has been demonstrated that so long as the objective of the network is to produce imperceptible adversarial noise then the use of a discriminator is an unnecessary computational burden. 
Our experiments also show that the model can be reduced in size substantially through knowledge distillation without an appreciable loss in performance.

We acknowledge that if adversarial image generation for face recognition such as we propose here were to become popular, then face recognition algorithms themselves could pivot and be trained on adversarial images rather than on unaltered images, and so the ``cold war” between face recognition and deception or polluting the models, would continue. Also, as part of our future work
it would be interesting to add multiple white-box and black-box detection models and to perform diversity experiments on several different datasets of face recognition.

\subsection{Acknowledgements} 
\noindent 
This work was part-funded by Science Foundation Ireland through grant number (SFI/12/RC/2289\_P2).



\small

\bibliographystyle{plain} 
\bibliography{bibfile.bib} 


\begin{biography}
\noindent 
Andrew Merrigan obtained an M.Sc. in Data Analytics from Dublin City University in 2021 and is a senior software engineer working for Blockdaemon Limited in the development of a multi-chain API for accessing blockchain data with a specialism in blockchain technology.
\\
\\
Alan Smeaton is Professor of Computing at Dublin City University. He received his PhD from University College Dublin and is an elected member of the Royal Irish Academy and a winner of the Academy's Gold Medal in Engineering Sciences, an award given  to individuals who have made an    outstanding scholarly contribution in their fields.  He is a Fellow of IEEE and a Principal Fellow of AdvanceHE.
\end{biography}

\end{document}